\documentclass[11pt,a4paper]{article}
\usepackage[hyperref]{emnlp2020}
\usepackage{times}
\usepackage{latexsym}

\usepackage{amsmath,amssymb}
\usepackage{color,soul}
\usepackage{enumitem} 
\usepackage{multirow}
\usepackage{graphicx}
\usepackage{subcaption}
\usepackage{longtable}

\usepackage{microtype}

\aclfinalcopy 

\title{Semantic coordinates analysis reveals language changes in the AI field}

\author{Zining Zhu$^{1,2}$, Yang Xu$^{1}$, Frank Rudzicz$^{3,4,1,2}$ \\
$^1$ University of Toronto, $^2$ Vector Institute, $^3$ Surgical Safety Technologies \\
$^4$ Li Ka Shing Knowledge Institute, St Michael's Hospital \\
\texttt{\{zining,yangxu,frank\}@cs.toronto.edu}
}

\date{}

\begin{document}
\maketitle
\begin{abstract}
Semantic shifts can reflect changes in beliefs across hundreds of years, but it is less clear whether trends in fast-changing communities across a short time can be detected.
We propose {\em semantic coordinates analysis}, a method based on semantic shifts, that reveals changes in language within publications of a field (we use AI as example) across a short time span. We use GloVe-style probability ratios to quantify the shifting directions and extents from multiple viewpoints.
We show that semantic coordinates analysis can detect shifts echoing changes of research interests (e.g., ``deep'' shifted further from ``rigorous'' to ``neural''), and developments of research activities (e,g., ``collaboration'' contains less ``competition'' than ``collaboration''), based on publications spanning as short as 10 years.

\end{abstract}

\section{Introduction}

The use of language is closely related to the semantics \citep{Sweetser1990,traugott2001regularity}. Language usage is deeply saturated with the beliefs and ideologies common in the communities \citep{levinson1983,dijk1977text,dijk1995discourse}. These beliefs are encoded in the discourse and represented in the context when people speak and write.

The semantics of words come from the contexts in which they occur \citep{frege1980foundations}. Distributed semantic models like word2vec \citep{Mikolov2013} and GloVe \citep{pennington2014glove} can represent the semantics of words by training vector representations using the word contexts.

Recently, several works have showed that probing such distributed semantic models can reveal subtleties in the common beliefs, biases, and perceptions about social classes, among communities \citep{caliskan183,kurita2019measuring,Kozlowski2019geometry}. Correspondingly, shifts in these bias would illustrate the development of the communities \citep{Garg2018} over long time spans (e.g., several decades). However, it is unknown whether semantic shifts are visible in {\em short} timespans and, if so, whether these expeditious shifts could reflect the developments of communities.

In this paper, we take the community of AI researchers as an example, because AI has undergone rapid developments in recent years, and there are abundant text-based resources, e.g., publications. With increased interest, the number of publications has grown almost exponentially \citep{sculley2018avoiding}. The rapid development of an academic field may incur varying consequences. For example,``misuse of language'' \citep{lipton2018troubling}, ``hypothesis after the results are known'' \citep{gencoglu2019hark,Kerr1998HARKing}, and concerns in peer reviewing quality \citep{sculley2018avoiding} are each common effects. 
Traditional approaches to analyzing trends in academic communities involved example-based evidence, which do not allow large-scale automatic analysis for academic articles. Can we analyze trends in AI using a data-driven approach?

Here, we answer this question by analyzing AI papers \emph{en masse} based on semantic shifts analysis, a method we refer to as \emph{semantic coordinates}.

Semantic coordinates describe the direction and extent of the meaning of a relatively ``fast-shifting'' \emph{target word} (e.g., ``deep'') shifting along a pair of relatively ``semantically stable'' \emph{coordinate words} (e.g., ``rigorous - neural''). For each time step, we use the log ratio of a GloVe-style co-occurrence probability to depict the position of the target word along the semantic coordinate.

Semantic coordinates analysis allows users to plug in ``target - coordinates'' sets conveniently. This paper presents 40 example sets -- each set contributing to an aspect -- illustrating the development trends in AI community across several categories:
\begin{itemize}[nosep]
    \item Research interests
    \item Semantic shifts correlated to word usages
    \item Changes in academic activities
\end{itemize}

\begin{figure*}[t]
    \centering
    \includegraphics[width=.32\linewidth]{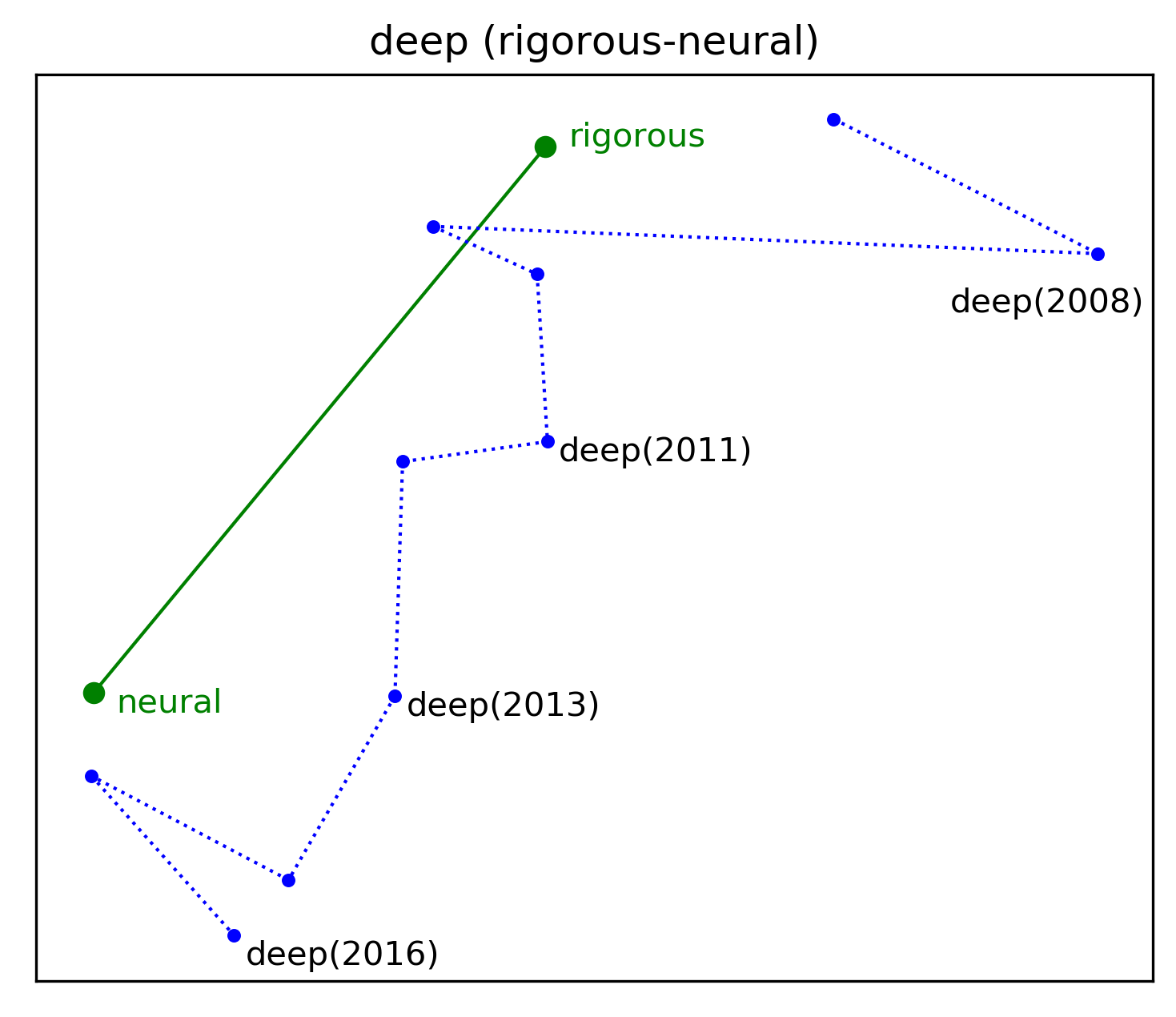}
    \includegraphics[width=.32\linewidth]{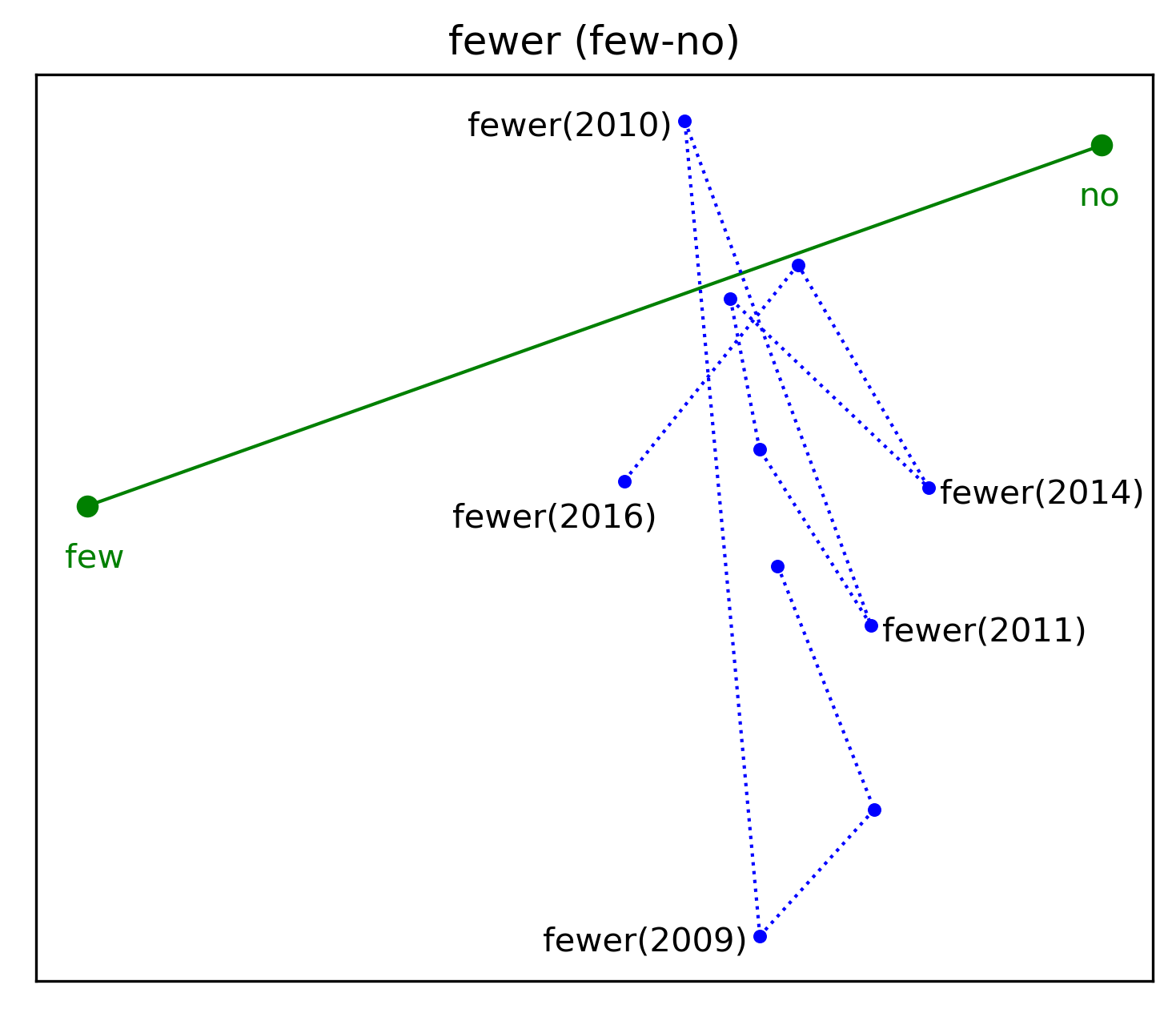}
    \includegraphics[width=.32\linewidth]{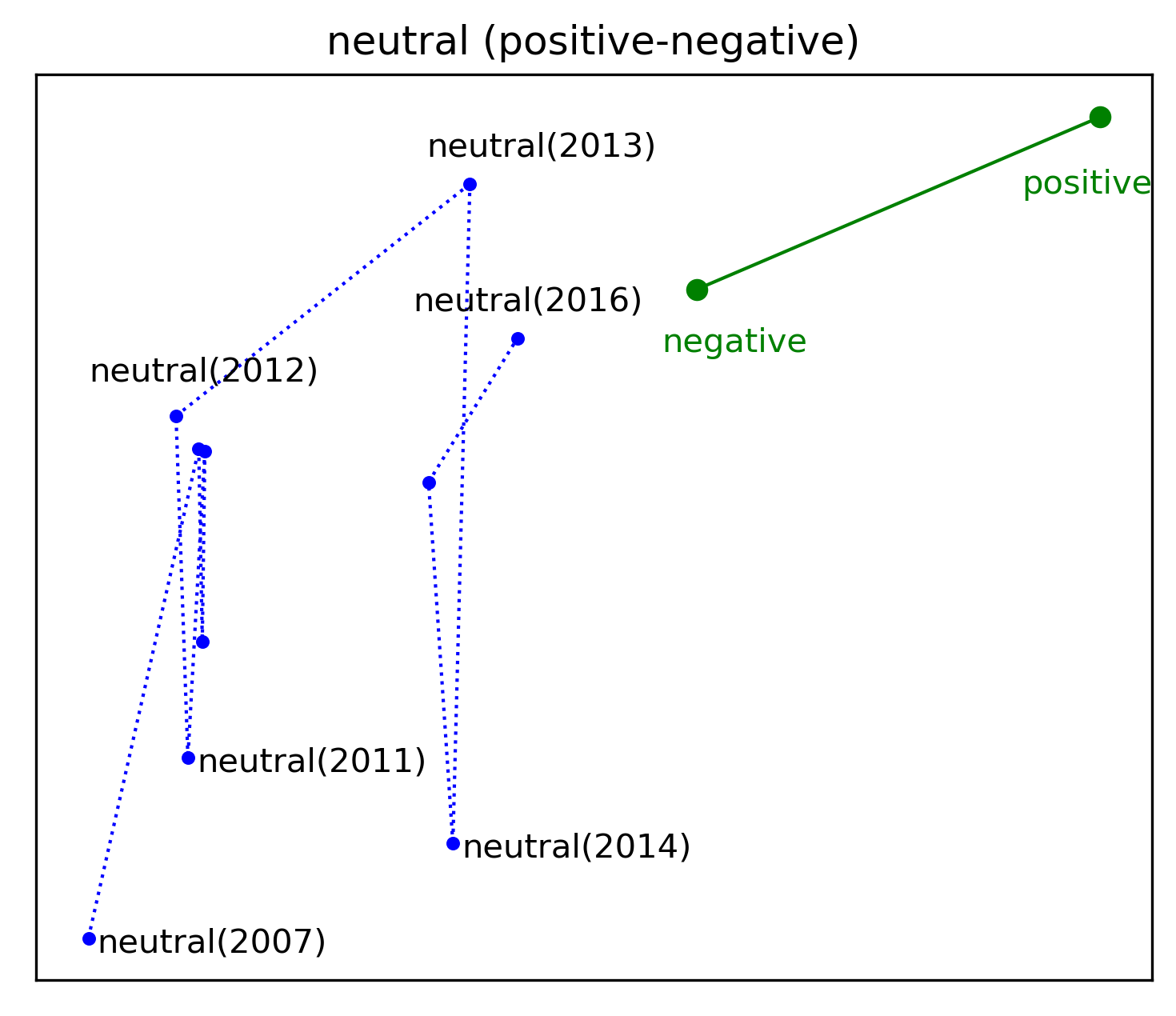}
    \caption{PCA visualization of three target words (``deep'', ``fewer'', ``neutral'') shifting on their corresponding semantic coordinates. From 2007 to 2016, ``deep'' clearly shifts away from ``rigorous'' and close to ``neural''. ``Fewer'' shifts almost perpendicularly to the ``few-no'' coordinate. ``Neutral'' shifts towards ``positive'' while still close to ``negative''. These types of shift are detected by the semantic coordinates.}
    \label{fig:pca}
\end{figure*}

Additionally, this paper presents ablation studies to address several confounding factors including the shifts of coordinate words. We find that time slicing and acceptance do not significantly change the shifts detected by semantic coordinates.

\section{Related work}
Our work is related to previous work in semantic shift detection, and describing development trends in academic fields.

\paragraph{Semantic shift detection}
\citet{hamilton2016cultural} compared semantic changes based on statistics of the local neighbors versus global embeddings of words, and showed that local measurement is more sensitive to semantic changes in nouns, whereas global measures \citep{Hamilton2016,kim-etal-2014-temporal,gulordava-baroni-2011-distributional} are more sensitive to changes in verbs.

There are many alternative methods to model changing semantics \citep{tahmasebi2018survey,schlechtweg-etal-2019-wind}. For example, works relying on Bayesian framework \citep{bamler2017dynamic,rudolph2018dynamic}, BERT \citep{hu-etal-2019-diachronic}, and graphical structures \citep{mitra-etal-2014-thats}. 
\citet{Dubossarsky2019} identified the shifting words at different times with distinct tokens, and trained a joint embedding space. Their approach focused on the {\em absolute} shifting of words within the joint embedding space, while our approach shows {\em relative} shifts along coordinates.

\citet{cook-stevenson-2010-automatically} used the difference of pointwise mutual information to measure the semantic polarity of words. They measured the amelioration and pejoration (i.e., words having more positive or negative meanings).
Later works analyzed semantic shifts along interpretable dimensions related to social science and digital humanities \citep{Garg2018,Kozlowski2019geometry}. \citet{hurtado-bodell-etal-2019-interpretable} used informative priors to incorporate domain knowledge into probabilistic semantic shifting models.

\paragraph{Semantic shifts and pragmatic usage}
\citet{breal1904essai} presented a taxonomy of semantic change, including restriction vs. expansion, pejoration vs. amelioration, and metaphor vs. metonomy. Subsequent literature consider the categories of semantic changes to have less definitive boundaries \citep{traugott2001regularity}.
\citet{Sweetser1990} identified the correlations between semantic changes, e.g., between modalities and language usages.
There are different theories between semantic change and pragmatic usage. For example, sequences of units used together undergo a ``chunking'' procedure and becomes a new complex unit \citep{bybee2010languageUsageCognition}. This procedure is domain agnostic, and exists in some of our findings as well. Another mechanism -- metaphorical mapping \citep{xu2017evolution} -- is related to the shift of word meanings from a source domain (e.g., ``grasp a fruit'') to a structurally similar target domain (e.g., ``grasp an idea''). The Conceptual Metaphor Theory proposed that people use metaphor to map concepts from a more concrete, physical domain to a more abstract, distant domain \citep{lakoff2008metaphors}.
All above works considered time spans on the order of hundreds of years. 

\paragraph{Trends in AI development}
The trends in academic development have been analyzed by numerous previous works. \citet{anderson2012comphistoryACL} used LDA-based algorithms to detect shifts in topic, and traced the interests of authors. 
\citet{webber-joshi-2012-discourse} traced the developments in discourse analysis.
\citet{hall2008Studying} studied the distributions of topics in computational linguistic conferences like COLING, ACL, and EMNLP.
However, none of these works analyzed the AI community in terms of the semantics and the usage of languages from word embedding approaches.

The development trends in AI development is also visible from articles discussing the scientific methods and how they can be improved.
\citet{forde2019scientific} described the scientific methods in physics, and related to the machine learning community.
\citet{gencoglu2019hark} mentioned the problem of ``hypothesizing after results are known'' in some machine learning work. 
\citet{sculley2018avoiding} mentioned problems in peer reviewing and suggested methods to improve peer reviewing qualities.
\citet{veysov2020sttcriticism} criticized various trends including leaderboard chasing and writing in obscure manners.

Besides the aforementioned research activities, the development of AI is accompanied by externalities including economic complexities \citep{mateos2018complex}, energy policies \citep{strubell-etal-2019-energy}, and climate change \citep{rolnick2019climate-change-ML}. Understanding the development in the AI community may help tackle the undesired impacts towards the socio-economic externalities.

\section{Methods}

\subsection{Diachronic word embedding}
To represent the semantics of words in different times, we use the diachronic word embedding similar to \citet{Hamilton2016}.

With each year as a ``time slice'', we train a word2vec embedding \citep{Mikolov2013} from arXiv articles published in this year. In addition, a ``base'' embedding is trained using articles from all years (i.e., from 2007 to 2016).

Following the approach of \citet{Hamilton2016}, we align the time sliced embeddings with a projection matrix. A time slicing embedding with vocabulary size $V$ at time $t$ is denoted with $W^{(t)}\in \mathbb{R}^{(V\times D)}$. A matrix $Q^{(t)}$ projects the time-sliced embedding matrix $W^{(t)}$ onto their representations in the ``base'' embedding space $W^{(t)}_{base}$. The projection matrix is optimized with the least squared loss: 
$$\min_{Q} ||W^{(t)}_{base} - W^{(t)}Q ||_F$$
where $|| \cdot ||_F$ denotes the Frobenius norm.
Applying a projection matrix corrects the corpus artefact discrepancy between different time-sliced portions of the corpus. 
Note that \citet{Hamilton2016} used the orthogonal Procrustes algorithm, whereas we relax the $Q^TQ=I$ constraint. This relaxation does not change all but two (i.e., $94.1\%$) of the trend-prediction results of the shifted words, but allows both translations and rotations between embedding spaces. This helps anneal the corpus artefact differences between time slices in academic publications.

As a sidenote, we consider diachronic word2vec more suitable for semantic coordiantes analysis than Temporal Referencing \citep{Dubossarsky2019}, for two reasons. First, TempRef requires designation of most words as non-shifting. This restricts the expressiveness of semantic models, especially when we do not know a large set of shifting words \emph{a priori}. Second, even if we successfully found a sufficiently large set of shifting words, TempRef-derived semantic coordinates analysis detects the ``global'' shifts, whereas we are more interested in the ``relative'' shifts, when finding the correlations between semantic changes and language usage. We focus on the diachronic word2vec henceforth.

We use a vocabulary of size $V=5000$ when training the projection matrix, but used all words occurring more than $f_{min}=3$ times when training word embedding. This would capture richer information encoded in those less frequent words, which otherwise would be collapsed onto \texttt{unk}.

\subsection{Semantic coordinates}
Now that we have $T$ embedding vectors for each word $w$, aligned onto the base embedding space. Our \emph{semantic coordinates} approach to analyze the directions and extents of semantic shifts consists of three steps, as will be described below.
\begin{enumerate}[nosep]
    \item Identification of target-coordinate,
    \item Computation of semantic coordinate positions,
    \item Computation of shifting trends.
\end{enumerate}

\paragraph{The target-coordinates structure} 
The ``target-coordinates'' structure is the core building block of semantic coordinates. A ``target-coordinates'' structure consists of a target word and a pair of \emph{coordinate words}. We assume the target-word experiences semantic shifts, while the coordinate words do not. Therefore, each pair of coordinates define a 1-d ``semantic axis'' allowing us to analyze the direction and extent along which the target words shift. 

Note that there could be multiple coordinate pairs per target word, each quantifying one latent dimension of semantic change. By identifying target-coordinates structures, we effectively project the semantic shifts onto interpretable dimensions allowing fine-grained analysis.

\paragraph{Identification of target-coordinates}
In semantic coordinates analysis, users can pick target words and corresponding coordinates of their interest. In this paper, we collect target words and coordinates using a combination of manual and semi-automatic approaches.

For the manual part, we include some instances reflecting whether some descriptor words experience semantic shifts\footnote{These 10 instances are: consider (certain, guess), novel (new, good), better (good, superior), improve (good, superior), first (new, good), early (new, good), extensive (lot, experiments), limitation (drawback, future), simple (obvious, method), fewer (few, no).}. For example, we are interested in whether ``novel'' means more ``good'' than ``new''. If all these words shifted to the more salient side, then authors tend to make larger claims. In experients, we observed shifting to the opposite directions.

The bulk of the target-coordinate sets we collect consists of words found in the semi-automatic approach. Following is a brief description.

We consider the most frequent 1,000 words as candidates for both the target and the coordinates. While less frequent words are more likely to experience semantic shifts \citep{gulordava-baroni-2011-distributional,Hamilton2016}, the shifts of infrequent words are more likely to be caused by corpus artefact differences instead of semantic changes, hence are less helpful for describing the linguistic trends of a community.

Among the more frequent 1000 words, we filter out the shifting words with a heuristic. Since the words shifting at the fastest ``speed'' in semantic space would have different neighbors (regardless of whether their neighbors have stable semantics), those words with the most changing neighbors should correspond to those with more salient semantic shifts. Therefore, we pick target words from the frequent words with at least one changing neighbor (i.e., with neighbors varying across the $T$ time slices). 

\paragraph{Computation of semantic coordinate positions}
With the diachronic embedding and the target-coordinates in place, we proceed with quantifying the shift of each target word $w$ along a semantic coordinate (denoted by its two coordinate words $(c_1, c_2)$). We use the ratio of co-occurrence probability to represent the position on the semantic coordinate.
\begin{equation}
d_{rel}^{(t)} = \text{log}\frac{P(w^{(t)}\,|\,c_1^{(t)})}{P(w^{(t)}\,|\,c_2^{(t)})} 
\label{eq:likelihood_ratio}
\end{equation}
This definition follows the intuition of \citet{pennington2014glove} and \citet{Ethayarajh2019}. They considered ``if $P(w\,|\,a)/P(w\,|\,b)=P(w\,|\,x)/P(w\,|\,y)$ for all $w$, then $a$ to $b$ is like $x$ to $y$''. When considering one target word $w$ at different time steps, such an analogy could be worded as ``if $c_1$ to $c_2$ (from the viewpoint of $w$) at time $t$ is like $c_1$ to $c_2$ (from the viewpoint of $w$) at time $t+1$, then the semantics of $w$ does not shift along the $(c_1 - c_2)$ coordinate''.

In addition, we take the log of probability ratio to demonstrate the position on semantic coordinate on $(-\infty, \infty)$ range (instead of the $(0, \infty)$ range). At time $t$, $d_{rel}^{(t)}>0$ indicates the target word $w$ being closer to $c_1$.

How can we represent the co-occurrence probability? We follow the distributional hypothesis \citep{harris1954distributional}, which claimed that those words with similar semantics have higher co-occurrence probabilities. Therefore, we use the cosine similarity of word embedding to represent these probabilities.
\begin{equation}
    d_{rel}^{(t)} = \text{log}\frac{\langle v_w^{(t)}, v_{c_1}^{(t)}\rangle}{\langle v_w^{(t)}, v_{c_2}^{(t)}\rangle},
\label{eq:reldist}
\end{equation}
where $v_{w}^{(t)}$ refers to the diachronic word embedding of word $w$ at time step $t$, and the cosine similarity of two vectors $v_1$ and $v_2$ is:
\begin{equation}
    \langle v_1, v_2 \rangle = \frac{v_1 \cdot v_2}{||v_1|| \cdot ||v_2||}
\label{eq:cos_dist}
\end{equation}

\begin{figure*}
    \centering
    \includegraphics[width=\linewidth]{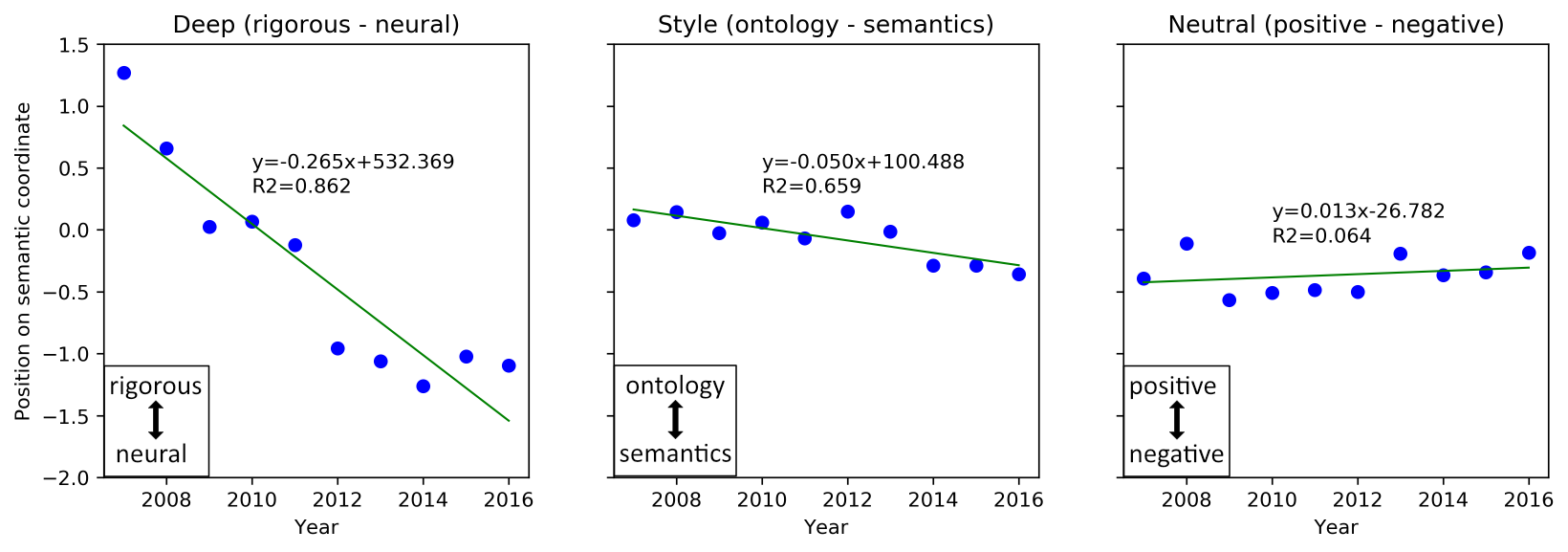}
    \caption{Plots showing the positions on semantic coordinates ($d_{rel}^{(t)}=\text{log}P(w\, | \, c_1) - \text{log}P(w\, |\, c_2 )$). 
    ``Deep'' clearly bears towards ``neural'', but much less shifting are detected with the other two examples.}
    \label{fig:trend_plot_1}
\end{figure*}

\paragraph{Trends of shifting}
Now we have a sequence of positions $\{ d_{rel}^{(t)} \}_{t=1}^{T}$ along semantic coordinates. To find the shifting direction, we fit a linear regression for each ``target-coordinates'' structure using least squared minimization, 
\begin{equation}
    k, b = \min_{k, b} | kt+b - d_{rel}^{(t)} |^2.
\label{eq:linear_fit}
\end{equation}
The sign of the slope $k$ indicates the direction, and the magnitude $|k|$ indicates the speed of the relative shift of the target word along its semantic coordinates.
Figure \ref{fig:trend_plot_1} and \ref{fig:trend_plot_2} show some examples.

\subsection{Implementation}
We use gensim \citep{gensim_rehurek_lrec} to train word2vec embeddings, and scikit-learn \citep{scikit-learn} to train the projection matrices.  In the process to train word embeddings, words occurring fewer than $f_{min}=3$ times are ignored. All embeddings have dimension $D=100$. For other hyperparameters, we rely on the default implementation of gensim. To increase reliability, we repeat training of word embeddings 4 times using distinct random seeds in each configuration, and average the slope to determine the final shifting direction.

\section{Data}
We use the AAPR dataset \citep{Yang2018AAPR}, consisting of 37,464 arXiv articles related to AI from 2007 to 2016. Approximately half of these (19,143) have been accepted for publication elsewhere. The accepted venues include major conferences and journals such as EMNLP, AAAI, ACL, Nature, and many workshops (e.g., MLMMI at ICML, MetaSel at ECAI, and NUT at COLING). 
As shown in Supplementary Material
, in general, there are increasing number of papers each year.
Table \ref{tab:datasets-comparison} shows a comparison of AAPR against two frequently used corpora in detecting semantic shifts. AAPR contains a comparable number of tokens with COHA \citep{COHA-dataset}, but over a much shorter timespan. 
\begin{table}[h]
    \centering
    \begin{tabular}{l c c}
    \hline 
         Dataset & Tokens & Time span \\ \hline 
         AAPR & $4.34 \times 10^8$ & 2007-2016 \\
         COHA & $4.06 \times 10^8$ & 1810s-2000s \\
         Google Ngram & $9.51\times 10^{10}$ & 1800s-2000s\\ \hline 
    \end{tabular}
    \caption{A comparison between the AAPR \citep{Yang2018AAPR}, COHA \citep{COHA-dataset}, and Google Ngram \citep{michel2011GoogleBooks} (the unigram portion)}
    \label{tab:datasets-comparison}
\end{table}

We preprocess the latex source codes by removing \LaTeX commands and equations. For the articles, we concatenate all sections in the article body including the abstract, experiment, discussion, and conclusion sections.

\begin{figure*}
    \centering
    \includegraphics[width=\linewidth]{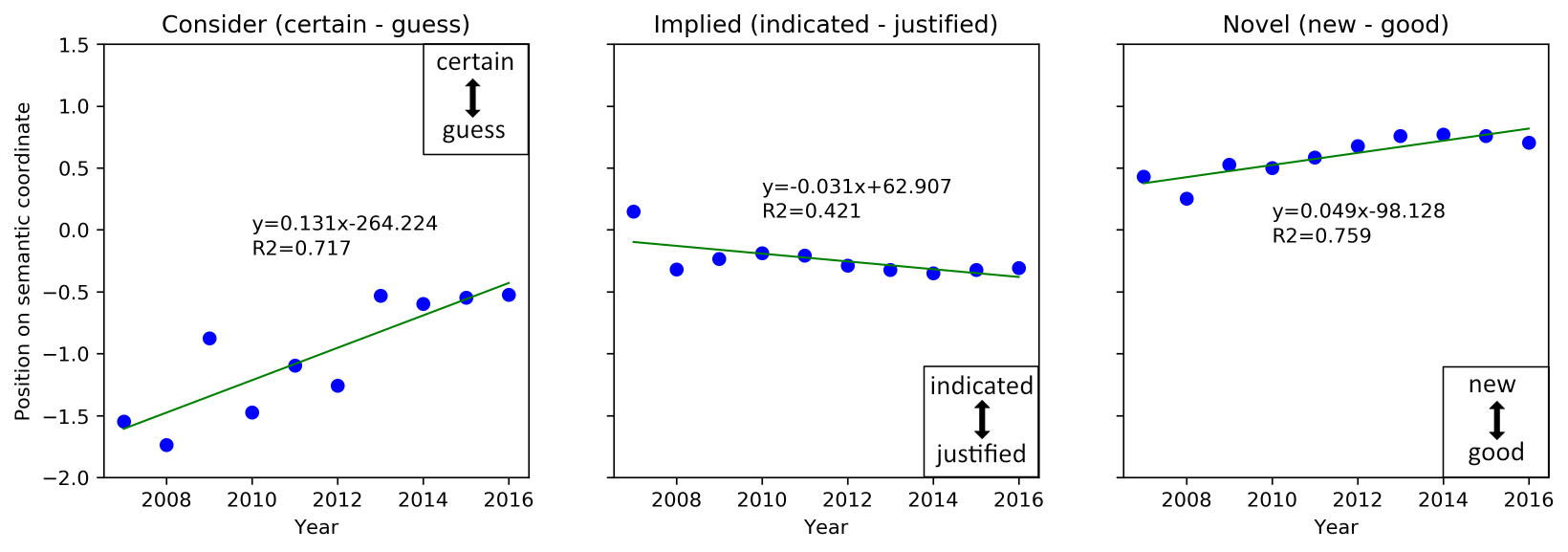}
    \caption{Three examples of semantic shifts showing that semantic coordinates preserve global proximity while following relative trends. A positive value indicates the proximity of the target word (``consider'', ``implied'', and ``novel'') to the former coordinate word (``certain'', ``indicated'' and ``new'') at the corresponding time step.}
    \label{fig:trend_plot_2}
\end{figure*}

\section{Results}

The semantic coordinates analysis reveals various developmental trends in AI. In this section, we roughly divide the shifting results of ``target-coordinates'' into the following categories. We include the full table of ``shift from'' and ``shift to'' coordinates of the target words, together with the slopes of linear fitting, in Appendix (Table \ref{tab:results}). In this section, we divide the results into four categories, i.e., shifts in research interests, shifts correlated to usage changes, shifts in academic activities, and the ``no detectable shifts''. Note that the shifting categories are not strictly exclusive. For each category, we present some examples and posit some possible reasons of these shifts. 

\paragraph{Shifts in research interests} This category includes some of the most obvious examples.

As shown in Figure \ref{fig:trend_plot_1}, ``deep'' bears towards ``neural'' on the ``rigorous-neural'' scale, as indicated by the negative slope in the linear fitting. Specifically, there is a ``stepping down'' in the year 2012, when AlexNet \citep{krizhevsky2012imagenet} improved on the ImageNet classification state-of-the-art by a significant margin.

The term ``learn'' would be related to ``know'' in the cognitive sense, whereas it would be relevant to ``predict'' when it describes the optimization procedure of parameterized algorithms. This shift away from its cognitive sense is also reflected on the ``receive-generate'' scale.

As another example, the word ``style'' gradually becomes more related to the semantics of, e.g., images and texts, compared to ontology, as shown in Figure \ref{fig:trend_plot_1}.
Unlike the ``deep'' (rigorous-neural) example, the changing point seems to occur around 2012. 
If we trace back to the most highly cited publications around that time, we see some landmark papers in semantic segmentation with convolutional neural networks (e.g., \citet{girshick2014RCNN,long2015fully}). We can also find precursors to neural image style transfer \citep{gatys2015neural}. The occurrence of these papers illustrates the shift in the meaning of ``style'' towards ``semantic''. Note that this shift, as reflected by the semantics in publications from the AI community, precedes the occurrence of these highly cited papers by 2-3 years.

For additional examples in shifting research interests, we include word cloud plots of representative conferences (AAAI, NeurIPS, ACL, and ICASSP) in the Supplementary Material.

\paragraph{Shifts correlated to usage changes} 
This category includes the changes in semantics reflecting either usage or assumptions.

The first three examples in the ``language usage'' category in Table 1 of the Supplementary Material (``novel'', ``first'', ``early'') shifted further away from ``good'' and closer to ``new''. This pejoration procedure could be easily confused with the ``language misuse'' trend mentioned by \citet{lipton2018troubling}. Instead of attributing to ``overclaiming'' and using ``more salient'' words intentionally, the observations could be alternatively explained with pejoration of qualifier words.

Are these pejorations caused by authors using these more salient words (so the co-occurrence patterns are reflected on word embeddings), or do the semantic changes motivate change in pragmatic usage (so the authors could present the same overall meanings)?
Unfortunately, semantic coordinates analysis does not yet have the counterfactual tools to identify the causality required to answer this question.

We also observe shifts in modality. For example, ``implied'' went from the more deontic senses (``indicated'' / ``influenced'') to the more epistemic senses (``justified'' / ``imposed'').

A noteworthy example is ``neutral''. As shown in Figure \ref{fig:pca}, although it shifts towards ``positive'', the semantics remains closer to ``negative'', and the slope of its shift is smaller than $90\%$ (36 out of 40) of all target words. Considering the $\langle \text{ratio},\text{positive}\rangle / \langle \text{ratio},\text{negative} \rangle$ ratio values are always negative for all years (as shown in Figure \ref{fig:trend_plot_1}), this is an example of semantic coordinates analysis taking the absolute positions into consideration.

\paragraph{Shifts in academic activities}
Semantic coordinates can also capture subtle changes in academic activities. These changes show early signs of metaphorical mappings \citep{lakoff2008metaphors}, with the two coordinate words indicating slightly different contexts in which the target word is used. Although the AI community is developing rapidly, the changes in contexts are less than the discrepancy between source and target domains considered by the metaphorical mapping literature (e.g., \citet{anderson2016mapping} considered metaphorical mappings in English across over 1000 years). Regardless, the trends of metaphorical mapping are visible throughout the ten year span.
Some examples are elaborated as follows.

``Collaboration'' shifted away from ``competition'' and closer to ``communication''. This shift appears very interesting, because we also observed an increasing number of authors per article, as shown in Figure \ref{fig:avg-author-per-paper}. These together might indicate a more benign environment for collaboration.

Writing proposals is another common academic activity. The coordinate words of ``proposal'' are related to the components (e.g., motivation vs. methodology) or the procedure (e.g., prototype vs. pipeline) or writing proposals \citep{porter2007proposals,deitz2016supporting}. The semantics of ``proposal'' shifts towards the latter of the aforementioned pairs, indicating some shifts in the relative emphasis. 

``Assessment'' was frequently used to evaluate the quality of algorithms or datasets \citep{Rodriguez-Galiano2012,kubat1997addressing}. ``Assessment'' is now closely related to prediction of performance on designated tasks \citep{wang2015smartgpa}. In machine learning for health. To assess, e.g., the health status of a patient involves extracting visual or audio features, and running trained models to predict the likelihood of impairment \citep{Zhou2016,Ghassemi2018review}. Note that the ``assess the algorithm'' meaning is also widely used nowadays, but many subjects assessed involve the machine learning models -- those with prediction abilities \citep{irvin2019chexpert}.

\paragraph{Global vs. relative changes} Up till now, we have mentioned only the local information (i.e., the shifting along the semantic coordinates). As Figure \ref{fig:trend_plot_2} shows, the semantic coordinate positions also encode global proximity information. 
For example, ``novel'' remains closer to ``new'', as reflected by the positive values on the ``new - good'' coordinate. 

An example for capturing both global and relative semantic changes is ``consider'' along the ``certain - guess'' coordinate. Although ``consider'' shifted rapidly towards ``certain'', the portion of ``guess'' remains dominant, as reflected by the negative values on Figure \ref{fig:trend_plot_2}.

\paragraph{No detectable change} We designate four target-coordinates structures into this ``no shift'' category, since the directions of shifting can not be stably reflected on the semantic coordinates. In the word embedding with random seed $13$, ``proposal'' (descriptor, finding) flipped sign in a 1-year time slice.
In the ablation studies, when we add the Procrustes constraint on the diachronic embedding projection matrix or change the time slicing to 2-year, ``better'', ``fewer'', and ``proposal'' flip their slope signs. With a 3-year time slicing, ``fewer'', ``likewise'', and ``proposal'' flip signs. Figure \ref{fig:pca} presents that ``fewer'' shifts almost in orthogonal directions to the ``few-no'' coordinate.
These examples show the instability of these signs, and render the shifts of words in this category dubious.

\section{Discussion}
To qualify the semantic coordinate analysis, we present several ablation studies, addressing some potential confounds.

\paragraph{Coordinates are relatively stable} In previous analysis, we assumed the semantics of coordinate words to be stable. We define a stability score to describe the stability here. 

For a word $w$ at time $t$, let $N_t = \{ n_{1...K}^{(t)} \}$ be its $K$ nearest neighbors.
Among the union of these neighboring sets $N_1 \cup ... \cup N_T$, we count the occurrences of the $K$ most frequent words as $f_{1..K}$.
Then we define the stability score for this word $S^{(K)}(w)$ to be the mean of these occurrences: 
$$S^{(K)}(w) = \frac{1}{K} \sum_{k=1}^{K} f_k$$
If a word shifts greatly in the semantic space through time, then its neighbors would change much. The occurrences, $f$, of the most frequent ones in the union would be small, leading to a low $S(w)$ value, for different $K$ values.

As shown in Table \ref{tab:stability-score-t-tests}, there are significant differences for all of $S^{(5)}$, $S^{(10)}$, and $S^{(20)}$ between the target and coordinate words. This indicates that the coordinate words are more stable than the target words.
\begin{table}[h]
    \centering
    \begin{tabular}{c c c c}
    \hline 
         & $S^{(5)}$ & $S^{(10)}$ & $S^{(20)}$ \\ \hline
         Mean (target) & .62 & .65 & .67 \\ 
         Mean (coords) & .72 & .73 & .76 \\
         $p$ value & .0020 & .0022 & .0013 \\ \hline 
    \end{tabular}
    \caption{The mean and $p$ value (of Mann-Whitney test). All three stability scores show significant difference.}
    \label{tab:stability-score-t-tests}
\end{table}

\paragraph{Time slicing methods can be flexible}
In previous analysis, the time slicing for diachronic embedding is set as one year per slice (with 10 steps in total). While this allows granular analysis for trends, each time slice only has a fraction of documents. To check the effect of the time slicing scheme, we repeat semantic coordinate analysis on 2-year-slices\footnote{i.e., 5 time steps in total.} and 3-year-slices\footnote{2007-2010, 2011-2013, 2014-2016. Since earlier years have fewer papers, we let the first slice contain four years.}.

Although longer time slicing reduces the randomness in word embeddings, fewer points could reduce the confidence in interpreting the linear regression slope. Regardless, $90\%$ and $87.5\%$ of the slopes retain their polarities for 2-year-slices and 3-year-slices, respectively. Moreover, if we exclude words with unstable shifting directions (i.e., those in the last section of Table 
3 in Supplementary Material, then the portion of slopes preserving polarity would rise to $97.2\%$ and $94.4\%$ respectively. The high stability indicates that the impact of different time slicing methods is small.

\paragraph{Paper acceptance does not matter}
A potential confounder for language usage is paper acceptance. One might hypothesize that the accepted papers use words in a way that are different from those non-accepted papers, leading to different semantic shifts. We test this hypothesis by controlling the paper acceptance here.

The AAPR corpus could be divided into accepted (``AC'') and not accepted (``NAC'') sub-corpora, each consisting of approximately half of the articles. We train two embeddings with the same configurations as our diachronic word embeddings. On a Wilcoxon signed-rank test, the cosine similarity $\langle w, c_1 \rangle$ ($p=0.7536$) and $\langle w, c_2 \rangle$ ($p=0.8092$), and the co-occurrence ratio $\frac{\langle w, c_1 \rangle}{\langle w, c_2 \rangle}$ ($p=0.6638$) are not significantly different between the AC and the NAC embeddings.

\begin{figure}[t]
    \centering
    \includegraphics[width=\linewidth]{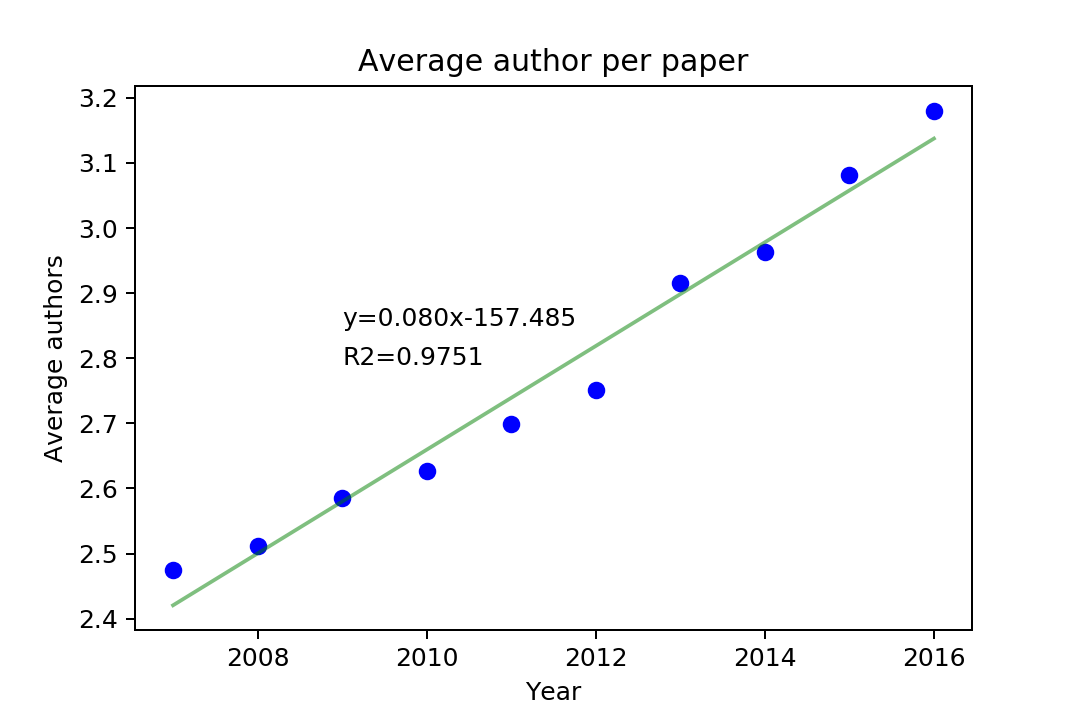}
    \caption{Average number of authors per paper. $18.2\%$ of the 2007 papers were written by a single author. This portion reduced to $11.2\%$ in 2016. This indicates that collaboration is increasingly prevalent.}
    \label{fig:avg-author-per-paper}
\end{figure}

\paragraph{Does language proficiency of author matter?}
Despite an increasing number of authors using English as a second language in recent years, we do not consider language proficiency as a causal factor in the semantic shifts. There are three reasons.
First, the language proficiency of an individual may improve with time. This improvement is hard to track without further author information.
Second, many authors from different language backgrounds collaborate on the same papers. Although they might start with different vocabularies and language usages, they tend to eventually reach a common ground \citep{stalnaker2002common-ground} during collaboration, sharing vocabularies and pragmatics in the paper. Considering the increasing inclination to collaborate (as shown in Figure \ref{fig:avg-author-per-paper}), this effect should counteract the differences in language proficiency. 
Third, academic conferences tend to encourage sharing of vocabulary and pragmatics. It is less likely that there are differences in language usages causing semantic gaps between groups of participants. 
In short, while more data is necessary to give a definitive answer, we find no evidence that language proficiency impacts the results of semantic coordinate analysis causally.

\section{Conclusion}
We introduce {\em semantic coordinates analysis} and show that it can detect semantic shifts within a fast-developing field like AI, using arXiv papers collected over as brief a period as a 10-year span. We show semantic coordinate analysis can detect shifts in research interests, changes in academic activities, and reflect correlations between semantic shifts and usage patterns. 
In the future, semantic coordinates analysis can be used to probe the development trends in more communities in a fine-grained but data-driven manner, based on arbitrary timestamped text.

\bibliography{emnlp2020}
\bibliographystyle{acl_natbib}

\appendix 
\newpage

\begin{table*}[!ht]
  \centering 
  \begin{tabular}{p{4cm} l p{3cm} l}
    \hline \hline
        Target word & Coordinate: from & Coordinate: to & Magnitude of slope \\ \hline 
        \multicolumn{4}{l}{\textit{Research interests}} \\ \hline
        deep & rigorous & neural & $0.2348 \pm 0.0180$ \\
        indicators & features & metrics & $0.0539 \pm 0.0076$ \\
        style & ontology & semantic & $0.0413 \pm 0.0069$ \\
        mnist & baseline & benchmark & $0.0254 \pm 0.0054$ \\
        mnist & train & test & $0.0104 \pm 0.0063$ \\
        learn & know & predict & $0.0210 \pm 0.0033$ \\
        learn & receive & generate & $0.0227 \pm 0.0036$ \\ \hline 

        \multicolumn{4}{l}{\textit{Language usages}} \\ \hline 
        novel & good & new & $0.0471 \pm 0.0020$\\
        first & good & new & $0.0360 \pm 0.0062$\\
        early & good & new & $0.0212 \pm 0.0063$ \\
        acceptable & honest & reasonable & $0.0659 \pm 0.0095$ \\
        neutral & negative & positive & $0.0182 \pm 0.0082$ \\
        extensive & lot & experiment & $0.0584 \pm 0.0081$\\
        complementary & equivalent & different & $0.0395 \pm 0.0099$ \\
        implied & indicated & justified & $0.0233 \pm 0.0050$ \\
        implied & influenced & imposed & $0.0466 \pm 0.0075$ \\
        basically & essential & actually & $0.0333 \pm 0.0042$ \\
        basically & quantitative & qualitative & $0.0188 \pm 0.0292$ \\
        totally & naturally & completely & $0.0481 \pm 0.0075$ \\ \hline 
 
        \multicolumn{4}{l}{\textit{Academic activities}} \\ \hline 
        consider & guess & certain & $0.1547 \pm 0.0360$ \\
        improve & superior & good & $0.0466 \pm 0.0043$ \\
        simple & method & obvious & $0.0072 \pm 0.0066$ \\
        limitation & future & drawback & $0.0791 \pm 0.0110$ \\ 
        supplementary & about & detailed & $0.0785 \pm 0.0048$ \\
        innovation & utilization & evolution & $0.1618 \pm 0.0104$ \\
        advice & followers & answer & $0.2022 \pm 0.0217$ \\
        advice & proposal & query & $0.1309 \pm 0.0116$ \\
        promise & concrete & possibility & $0.0734 \pm 0.0125$ \\
        proposal & benchmark & candidate & $0.0577 \pm 0.0035$ \\
        proposal & motivation & methodology & $0.0274 \pm 0.0035$ \\
        proposal & prototype & pipeline & $0.0699 \pm 0.0068$ \\
        review & comment & comprehensive & $0.0131 \pm 0.0068$ \\
        reporting & answering & measuring & $0.0117 \pm 0.0090$ \\
        assessment & validation & prediction & $0.0314 \pm 0.0051$ \\
        quantified & constrained & determined & $0.0856 \pm 0.0038$ \\
        collaboration & competition & communication & $0.0160 \pm 0.0075$ \\ \hline 

        \multicolumn{4}{l}{\textit{No detectable shift}}\\ \hline 
        better & \multicolumn{2}{c}{good, superior} & $0.0035 \pm 0.0015$ \\
        likewise & \multicolumn{2}{c}{then, thus} & $0.0040 \pm 0.0024$ \\
        proposal & \multicolumn{2}{c}{finding, descriptor} & $0.0068 \pm 0.0269$ \\
        fewer & \multicolumn{2}{c}{no, few} & $0.0071 \pm 0.0047$ \\
        \hline \hline
    \end{tabular}
    \caption{Target-coordinates shifting results, averaging from 4 independent experiments (with random seeds 0, 7, 13, 73 in training gensim word2vec embeddings).}
  \label{tab:results}
\end{table*}

\section{Full list of results}
We include the shifting directions and linear fitting slopes (averaged across 4 different random seeds) for all target-coordinates sets in Table \ref{tab:results}.

\section{Additional preprocessing details}
\paragraph{Why not lemmatize?} We consider lemmatized words to preserve lexical semantics and little morphological information. When training word embeddings, we want the word2vec model have larger capacity to capture the rich information. This is why we did not lemmatize words, or restrict to 5k vocabulary in training word embeddings. As is shown on nearest neighbor examples in Table \ref{tab:nearest_neighbors}, some words stemming from the same lemma (e.g., ``communication'' and ``communications'') are distributed in nearby locations in the semantic space, as they have similar nearest neighbors. However, when the word morphology possess heavier semantics (e.g., the past tense of ``implied''), the diachronic word2vec models assemble many verbs also in past tense to their neighborhood.

\section{Visualization of topic changes} 
\begin{figure*}
    \centering
    \begin{subfigure}{\textwidth}
        \includegraphics[width=.32\linewidth]{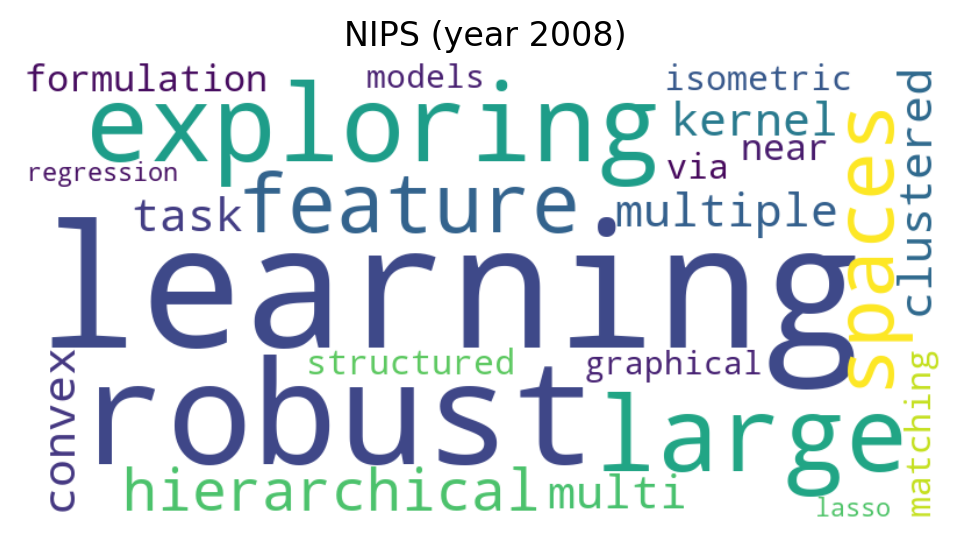}
        \includegraphics[width=.32\linewidth]{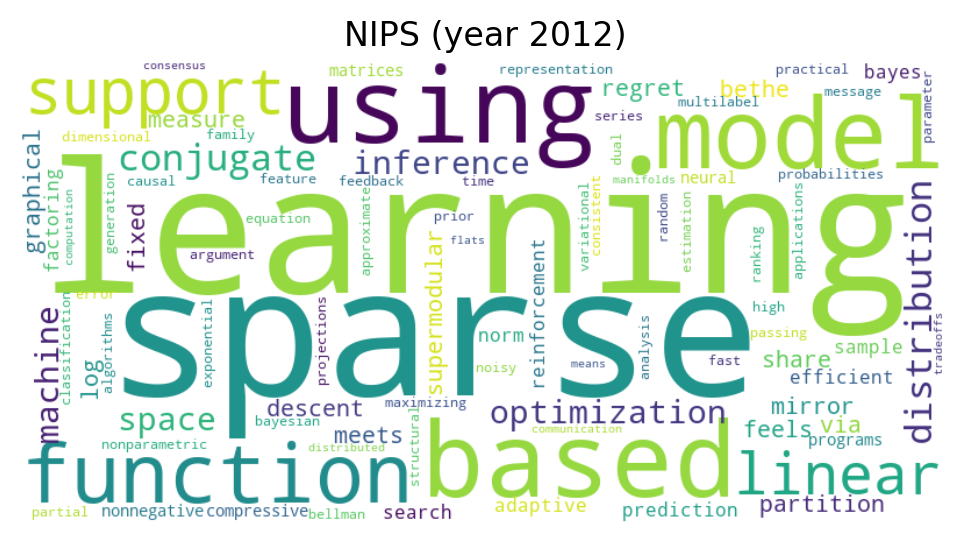}
        \includegraphics[width=.32\linewidth]{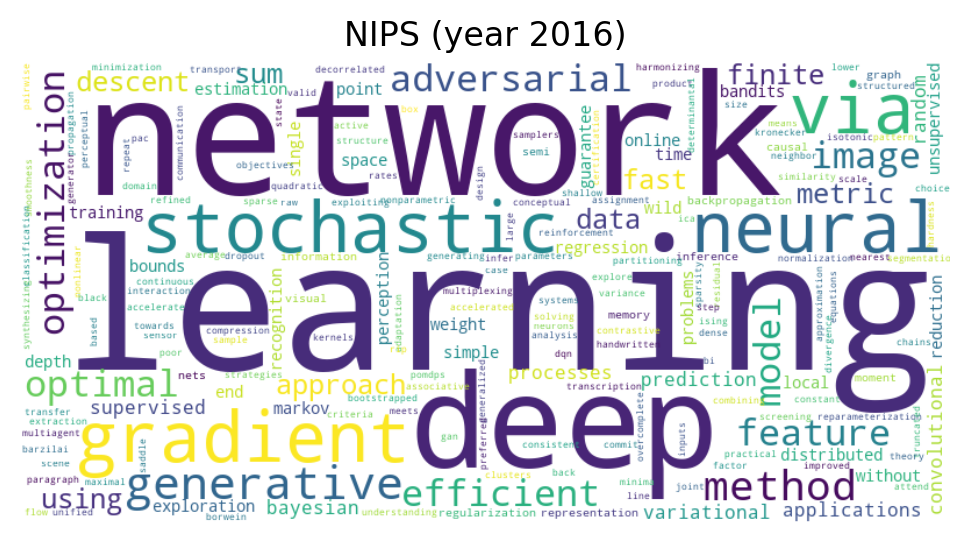}
        \caption{Wordcloud for abstracts in NeurIPS (formerly NIPS). The papers focus on learning. However, earlier papers are more interested in feature-based learning, while more recent research interests are focused on deep networks, gradients, and generative models.}
    \end{subfigure}
    \begin{subfigure}{\textwidth} 
        \includegraphics[width=.32\linewidth]{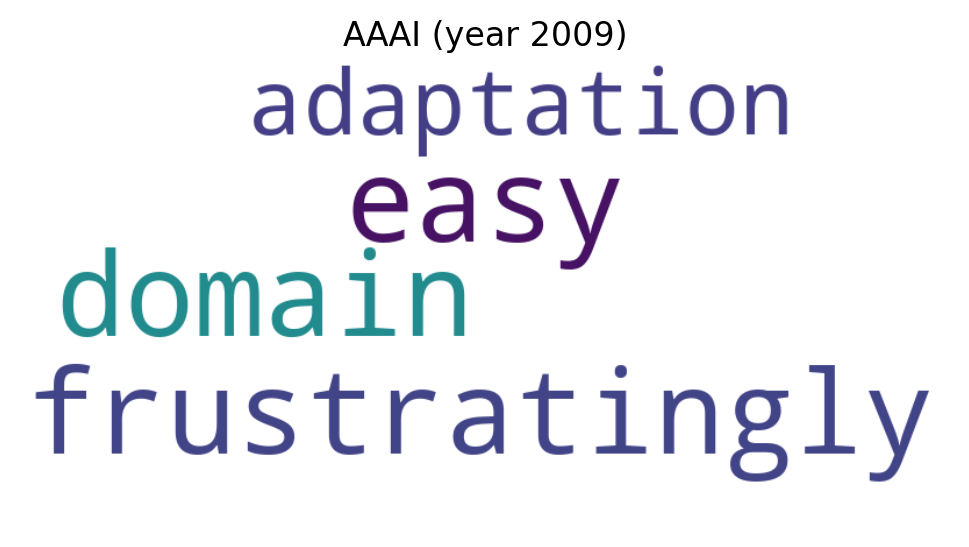}
        \includegraphics[width=.32\linewidth]{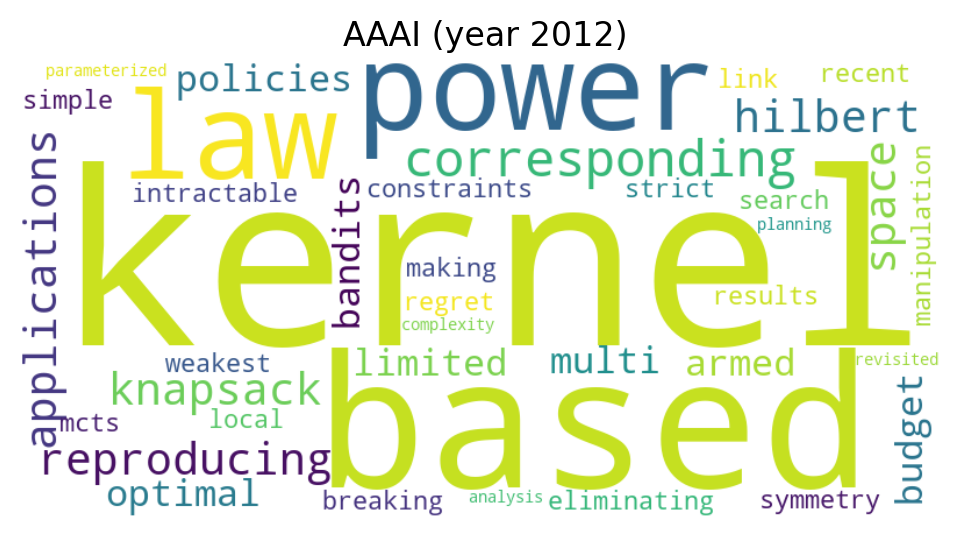}
        \includegraphics[width=.32\linewidth]{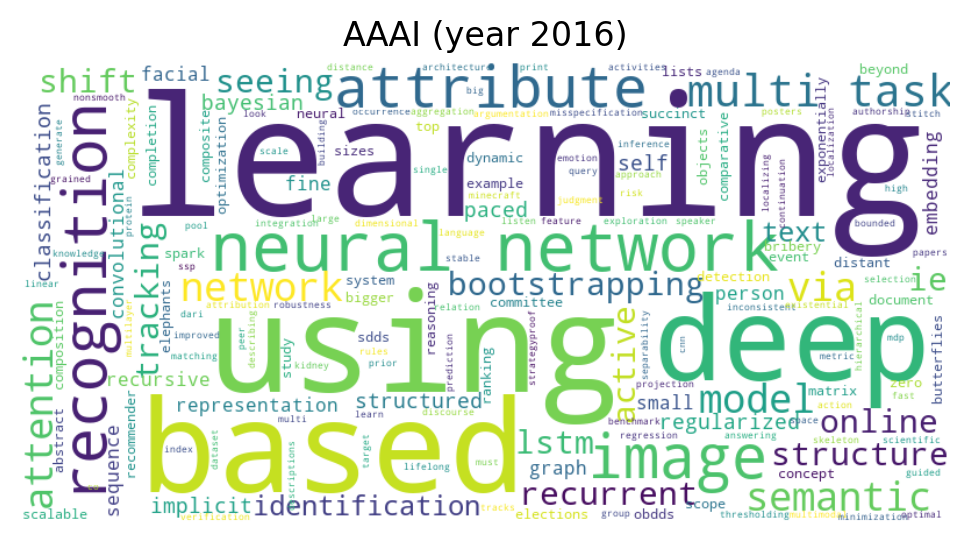}
        \caption{AAAI papers involved domain adaptation in 2009. In 2012, kernel-based methods became more popular. In 2016, neural network and deep learning took their places. Tasks including image recognition and multi-task learning are also mentioned frequently.}
    \end{subfigure}
    \begin{subfigure}{\textwidth}
        \includegraphics[width=.32\linewidth]{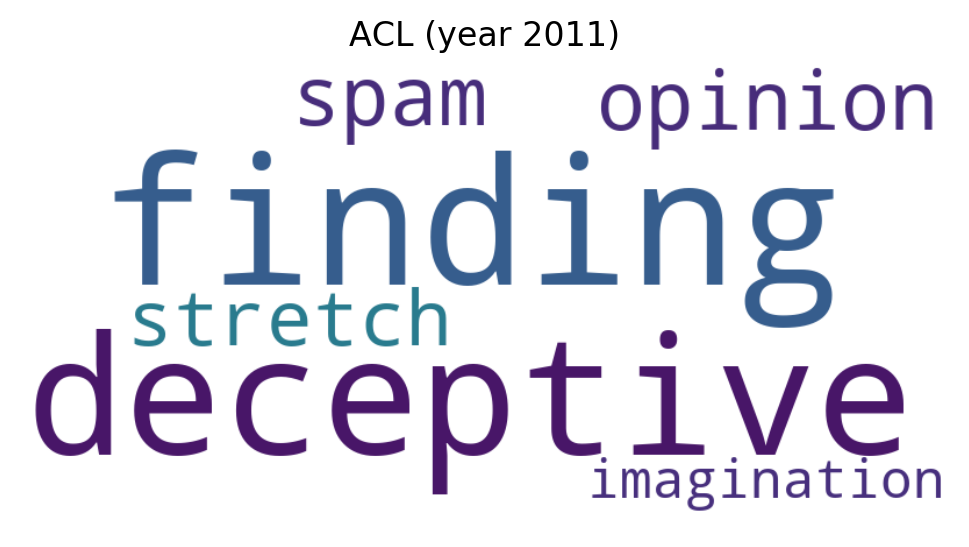}
        \includegraphics[width=.32\linewidth]{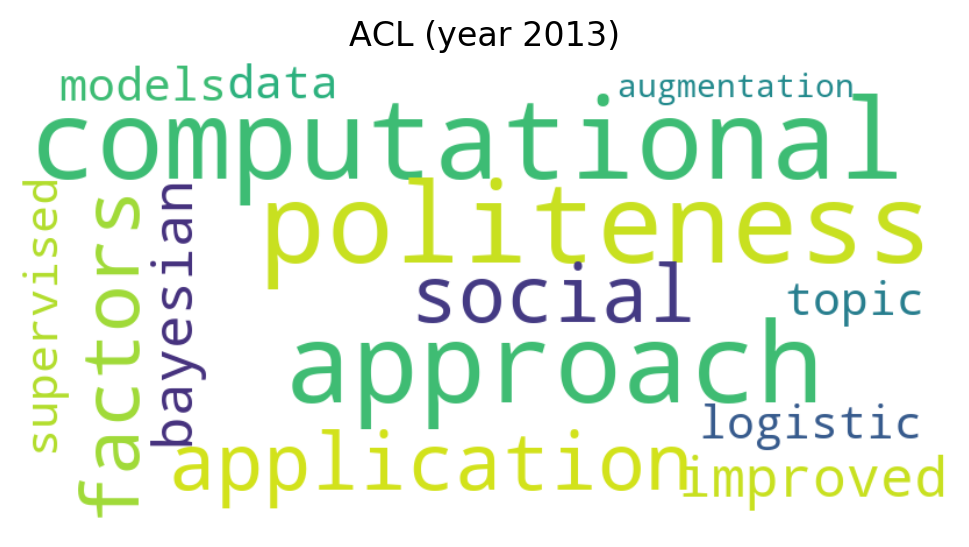}
        \includegraphics[width=.32\linewidth]{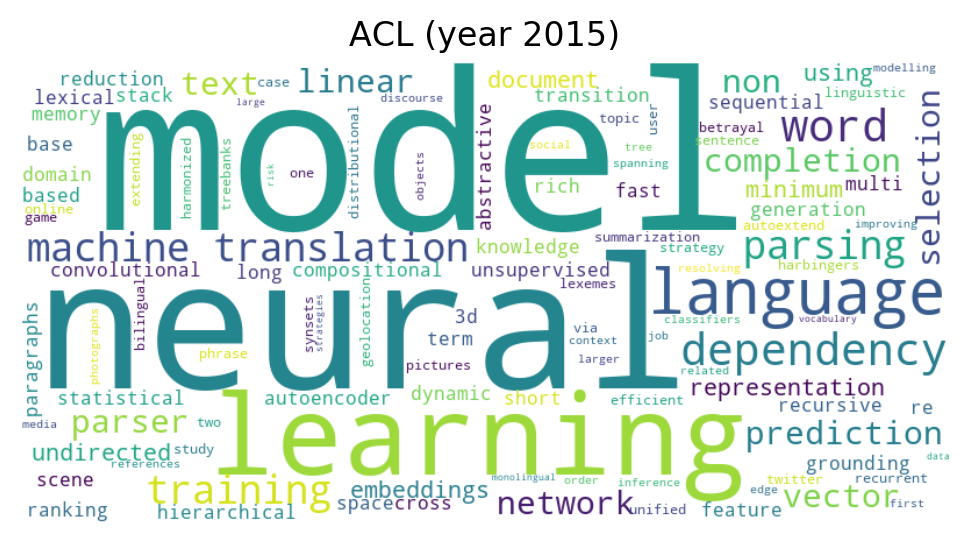}
        \caption{For the ACL meeting in 2011, popular topics included spam detection and opinion analysis. In 2013, numerous papers about social applications emerge. Keywords about machine learning including ``Bayesian'' and ``logistic'' emerged. In 2015, neural models became the most frequent terms, while topics including language modeling, machine translation, dependency parsing are mentioned a lot as well.}
    \end{subfigure}
    \begin{subfigure}{\textwidth}
        \includegraphics[width=.32\linewidth]{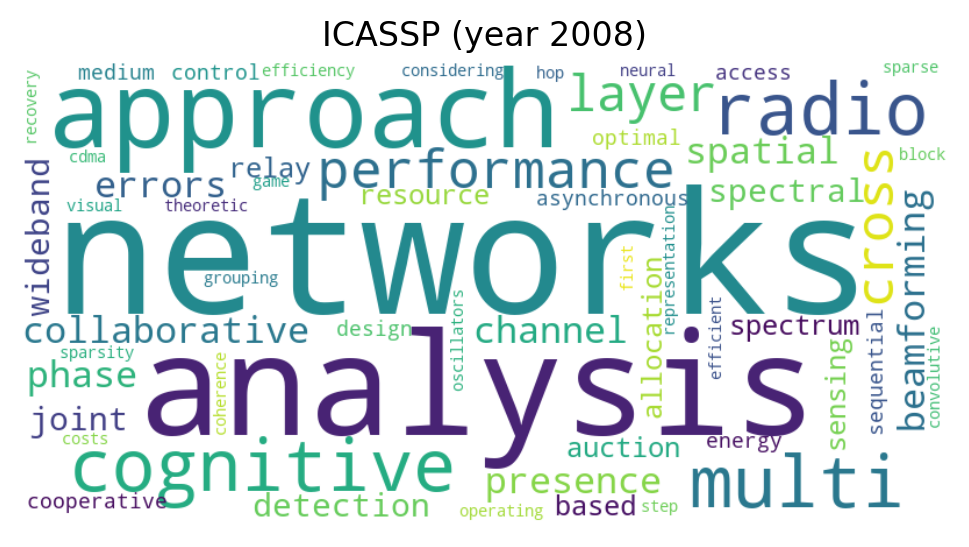}
        \includegraphics[width=.32\linewidth]{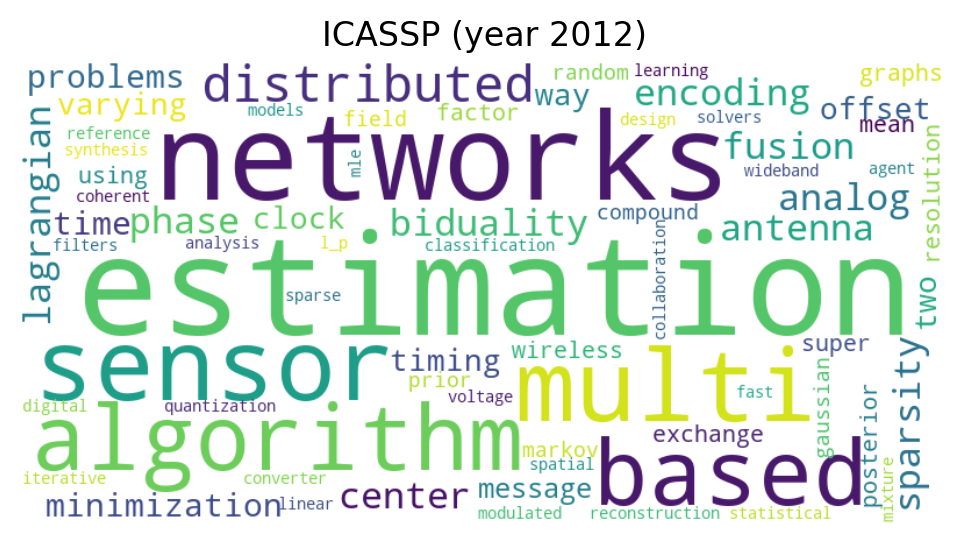}
        \includegraphics[width=.32\linewidth]{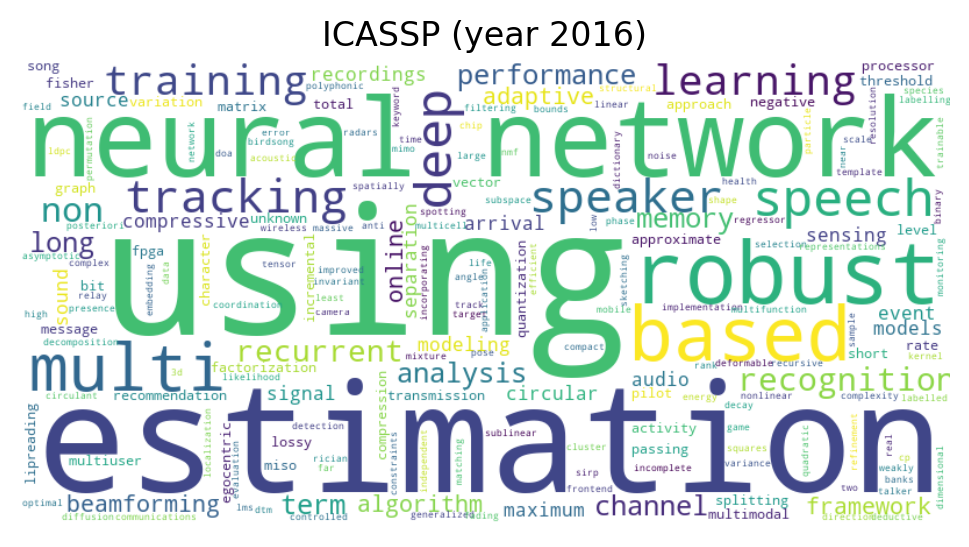}
        \caption{ICASSP abstracts in 2008 contained multiple mentions of ``networks'' together with ``analysis'' and ``approach''. In 2016, however, the term ``neural network'' and accompanying machine learning terminologies (e.g., ``estimation'') became more frequent.}
    \end{subfigure}
    \caption{Wordcloud visualizations showing the most frequent words on arXiv papers in AAPR belonging to NeurIPS, AAAI, ACL, and ICASSP respectively. Note that early conferences contain less articles on arXiv, resulting in more sparse wordcloud visualizations.}
    \label{fig:wordcloud_visualization}
\end{figure*}

We include wordcloud visualizations of the most frequent words in the abstracts of NeurIPS, AAAI, ACL, and ICASSP in Figure \ref{fig:wordcloud_visualization}.

\section{Examples of nearest neighbors}
We present some nearest neighbor of target words from 2007 to 2016 in the diachronic word2vec embedding space. In Table \ref{tab:nearest_neighbors} and Table \ref{tab:nearest_neighbors_2}, the \textul{underlined} words are coordinate words, and the \textbf{\textul{bold}} words are those that the target words shift towards. In each year, the nearest neighbors are displayed closest word first.
The coordinate words do not always occur in the nearest neighbors, but their locations on the tables give intuitive, qualitative visualizations of the semantic shifts.

\begin{table*}
    \centering
    \begin{tabular}{c c p{11cm}}
    \hline 
        Word & Year & Nearest neighbors in 5k vocab \\ \hline 
        \multirow{20}{*}{collaboration} & 2007 & infrastructure, internet, remote, commercial, personal, health, community, deployment, servers \\
        & 2008 & infrastructure, deployment, \textul{competition}, technologies, coordination, collaborative, cooperation, standards, opportunities \\
        & 2009 & https, security, business, addressing, \textul{competition}, researchers, attacks, management, infrastructure \\
        & 2010 & social, economic, organization, facebook, business, internet, transportation, \textul{competition}, brain \\
        & 2011 & social, political, facebook, transportation, adoption, financial, internet, economic, emergence \\
        & 2012 & adoption, cooperation, coordination, \textbf{\textul{communication}}, diversity, social, heterogeneity, communications, deployment \\
        & 2013 & social, business, scientists, political, scientific, internet, interactions, organization, infrastructure \\
        & 2014 & social, interaction, trade, interactions, facebook, transportation, connections, cooperation, financial \\
        & 2015 & social, interaction, facebook, coordination, cooperation, \textul{competition}, interactions, academic, trade \\
        & 2016 & \textbf{\textul{communication}}, communications, transportation, deployment, sharing, cooperation, coordination, discovery, cooperative \\ \hline 
        
        \multirow{20}{*}{implied} & 2007 & preserved, \textul{influenced}, true, \textul{indicated}, dominated, violated, absolute, replaced, covered \\
        & 2008 & satisfied, preserved, dominated, met, replaced, \textbf{\textul{justified}}, determined, multiplied, verified \\
        & 2009 & replaced, \textbf{\textul{justified}}, satisfied, violated, \textbf{\textul{imposed}}, dominated, multiplied, determined, specified \\
        & 2010 & replaced, preserved, dominated, \textbf{\textul{justified}}, \textbf{\textul{imposed}}, satisfied, affected, characterized, captured \\
        & 2011 & \textbf{\textul{justified}}, replaced, dominated, \textbf{\textul{imposed}}, satisfied, determined, characterized, preserved, \textul{influenced} \\
        & 2012 & dominated, quantified, preserved, \textbf{\textul{justified}}, affected, satisfied, replaced, violated, determined \\
        & 2013 & \textbf{\textul{justified}}, preserved, violated, \textbf{\textul{imposed}}, dominated, satisfied, replaced, induced, motivated \\
        & 2014 & \textbf{\textul{justified}}, preserved, satisfied, \textbf{\textul{imposed}}, replaced, violated, dominated, definable, induced \\
        & 2015 & \textbf{\textul{justified}}, \textbf{\textul{imposed}}, satisfied, preserved, replaced, violated, dominated, motivated, characterized \\
        & 2016 & \textbf{\textul{justified}}, \textbf{\textul{imposed}}, dominated, satisfied, characterized, induced, \textul{influenced}, preserved, motivated \\ \hline 
    \end{tabular}
    \caption{Showing semantic shifts using nearest neighbors. The \textul{underlined} words are the coordinate words we identified, and the \textbf{\textul{bold}} coordinate words are those that the target word shifts towards. Three ``target-coordinates'' are shown here: (collaboration, \textul{competition}, \textbf{\textul{communication}}), (implied, \textul{indicated}, \textbf{\textul{justified}}), and (implied, \textul{influenced}, \textbf{\textul{imposed}}).}
    \label{tab:nearest_neighbors}
\end{table*}

\begin{table*}
\centering
    \begin{tabular}{c c p{11cm}}
    \hline 
        Word & Year & Nearest neighbors in 5k vocab \\ \hline 
        \multirow{20}{*}{acceptable} & 2007 & \textul{honest}, violated, maximized, identical, decreased, minimized, delivered, guaranteed, admissible \\
        & 2008 & preferred, effective, sensitive, high, meaningful, costly, increased, attractive, accurate \\
        & 2009 & \textbf{\textul{reasonable}}, appropriate, meaningful, possible, required, desirable, good, inconsistent, intractable \\
        & 2010 & incorrect, \textbf{\textul{reasonable}}, optimal, accurate, infeasible, increased, necessary, required, excellent \\
        & 2011 & \textbf{\textul{reasonable}}, desirable, unlikely, good, excellent, infeasible, beneficial, high, interesting \\
        & 2012 & \textbf{\textul{reasonable}}, meaningful, good, significant, beneficial, enhanced, comparable, poor, effective \\
        & 2013 & \textbf{\textul{reasonable}}, good, effective, accurate, incorrect, appropriate, optimal, desirable, comparable \\
        & 2014 & \textbf{\textul{reasonable}}, effective, excellent, good, expected, preferred, increased, high, desirable \\
        & 2015 & \textbf{\textul{reasonable}}, effective, good, desirable, appropriate, beneficial, accurate, high, low \\
        & 2016 & \textbf{\textul{reasonable}}, effective, appropriate, good, optimal, attractive, desirable, expected, bad \\ \hline 
        
        \multirow{19}{*}{deep} & 2007 & \textul{rigorous}, shallow, nonetheless, certainly, nice, complicated, logics, weak, probabilistic \\
        & 2008 & linking, rich, strong, deeper, narrow, capturing, obstacles, lot, vast \\
        & 2009 & sophisticated, massive, flexible, traditional, powerful, deeper, simple, rich, texture \\
        & 2010 & deeper, rich, biological, multiscale, powerful, behavioral, shallow, promising, structural \\
        & 2011 & molecular, rapid, little, shallow, rich, deeper, powerful, realistic, multiscale \\
        & 2012 & \textbf{\textul{neural}}, unsupervised, deeper, biological, structured, visual, diverse, localized, promising \\
        & 2013 & \textbf{\textul{neural}}, shallow, convolutional, rnn, supervised, multilayer, unsupervised, deeper, feedforward \\
        & 2014 & \textbf{\textul{neural}}, supervised, cnn, unsupervised, shallow, convolutional, discriminative, generative, multilayer \\
        & 2015 & \textbf{\textul{neural}}, supervised, cnn, unsupervised, convolutional, recurrent, deeper, discriminative, shallow \\
        & 2016 & \textbf{\textul{neural}}, cnn, convolutional, deeper, supervised, unsupervised, recurrent, shallow, multiscale \\ \hline 
        
    \end{tabular}
    \caption{Additional examples showing semantic shifts using nearest neighbors. Two ``target-coordinates'' are shown here: (acceptable, \textul{honest}, \textbf{\textul{reasonable}}), (deep, \textul{rigorous}, \textbf{\textul{neural}}), and (implied, \textul{influenced}, \textbf{\textul{imposed}}).}
    \label{tab:nearest_neighbors_2}
\end{table*}
\end{document}